\documentclass[sigconf,table]{acmart}
\usepackage{listings}
\usepackage{csquotes}
\usepackage{subcaption}
\usepackage{multirow}
\usepackage{xcolor}
\usepackage{fancyhdr}

\AtBeginDocument{%
  }

\newenvironment{compactItemize}{\begin{list}{$\bullet$}
{\setlength{\topsep}{1mm}\setlength{\itemsep}{0.25mm}
\setlength{\parsep}{0.1mm}
\setlength{\itemindent}{0mm}\setlength{\partopsep}{0mm}
\setlength{\labelwidth}{15mm}
\setlength{\leftmargin}{4mm}}}{\end{list}}

\copyrightyear{2025}
\acmYear{2025}
\setcopyright{cc}
\setcctype{by}
\acmConference[WWW Companion '25]{Companion Proceedings of the ACM Web Conference 2025}{April 28-May 2, 2025}{Sydney, NSW, Australia}
\acmBooktitle{Companion Proceedings of the ACM Web Conference 2025 (WWW Companion '25), April 28-May 2, 2025, Sydney, NSW, Australia}
\acmDOI{10.1145/3701716.3715473}
\acmISBN{979-8-4007-1331-6/25/04}
\begin{document}

\fancypagestyle{firststyle}
{
   \fancyhf{}
   \fancyhead[C]{This is an authors' copy of the paper  to appear in  Companion Proceedings of the ACM Web Conference
2025 (WWW Companion '25).}
   \renewcommand{\headrulewidth}{0pt} 
}
\title[Personalizing LLMs using Retrieval Augmented Generation and Knowledge Graph]{Personalizing Large Language Models using Retrieval Augmented Generation and Knowledge Graph}
%


\author{Deeksha Prahlad}
\affiliation{%
 \institution{Arizona State University}
 \city{Tempe}
 \state{AZ}
 \country{United States}}
\email{dprahlad@asu.edu}

\author{Chanhee Lee}
\affiliation{%
 \institution{Arizona State University}
 \city{Tempe}
 \state{AZ}
 \country{United States}}
\email{chanheel@asu.edu}

\author{Dongha Kim}
\affiliation{%
 \institution{Arizona State University}
 \city{Tempe}
 \state{AZ}
 \country{United States}}
\email{dongha@asu.edu}

\author{Hokeun Kim}
\affiliation{%
 \institution{Arizona State University}
 \city{Tempe}
 \state{AZ}
 \country{United States}}
\email{hokeun@asu.edu}

\renewcommand{\shortauthors}{Deeksha Prahlad, Chanhee Lee, Dongha Kim, \& Hokeun Kim}

\begin{abstract}
The advent of large language models (LLMs) has allowed numerous applications, including the generation of queried responses, to be leveraged in chatbots and other conversational assistants.
Being trained on a plethora of data, LLMs often undergo high levels of over-fitting, resulting in the generation of extra and incorrect data, thus causing hallucinations in output generation. 
One of the root causes of such problems is the lack of timely, factual, and personalized information fed to the LLM.
In this paper, we propose an approach to address these problems by introducing retrieval augmented generation (RAG) using knowledge graphs (KGs) to assist the LLM in personalized response generation tailored to the users.
KGs have the advantage of storing continuously updated factual information in a structured way.
While our KGs can be used for a variety of frequently updated personal data, such as calendar, contact, and location data, we focus on calendar data in this paper.
Our experimental results show that our approach works significantly better in understanding personal information and generating accurate responses compared to the baseline LLMs using personal data as text inputs, with a moderate reduction in response time.
\end{abstract}




\maketitle
\thispagestyle{firststyle}

\section{Introduction}
Large language models (LLMs) are gaining popularity in various applications for text classification, text summarization, question answering, and sentiment analysis.
Large models like GPT-4~\cite{achiam2023gpt} have demonstrated their prowess in analyzing and generating code over standard tasks.
However, because these models lack new domain-specific knowledge, they suffer from hallucinations~\cite{ji2023towards, galitsky2024truth, perkovic2024hallucinations}.
Leveraging retrieval augmented generation (RAG) to minimize hallucinations has set a standard benchmark for domain-specific question-answer applications~\cite{10.5555/3495724.3496517}.
RAG increases the reliability of LLM results~\cite{hu2024prompt,borgeaud2022improving} by introducing accurate data sources, which also eliminates factual errors.


Knowledge graphs (KGs)~\cite{10.1145/3447772}, by definition, are based on graph databases, where the nodes represent entities of interest, and the edges represent the relations between the nodes. 
Prominent open-source KGs include Wikidata~\cite{wikiData}, Freebase~\cite{freebase}, and DBpedia~\cite{DBpedia}. The structure of the knowledge graph is designed to evolve, which makes it possible to have the latest information. KGs can be created using databases and graph databases~\cite{das2020issues} such as neo4j, ArangoDB, OrientDB, and Infinite Graph.
Query languages that are used to query KGs~\cite{angles2017foundations} include SPARQL, CYPHER, and Gremlin.

Various studies have integrated KGs with LLMs to enhance the accuracy of LLM-generated responses.
KGs provide factual knowledge to address LLMs' drawbacks, such as hallucinations and one notable approach addressing these drawbacks of LLMs and KGs is Pan \textit{et al.}~\cite{10387715}.
Yang \textit{et al.}~\cite{yang2024give} enhance LLMs with KGs to overcome LLMs' limitations in recalling facts while generating contents.

A growing body of research focuses on the exciting potential of LLM personalization, paving the way for its diverse and personalized applications.
Baek \textit{et al.}~\cite{baek2024knowledge} propose augmenting relevant context to LLMs using users' interaction histories targeting web search scenarios where LLMs are used as search engines, also exploring KGs to store data; still it is based on web information, incurring data privacy issues.
Shen \textit{el al.}~\cite{shen2024pmg} propose personalized multimodal generation (PMG) of LLMs by converting user behaviors to language models to understand user preferences.
Qin \textit{et al.}~\cite{qin2024enabling} have implemented a framework for selecting the most representative online data through a quality metric system for on-device LLM personalization.
Lee \textit{et al.}~\cite{lee2024work} sketch out ideas for the on-device RAG for personalization using KGs without concrete design.

In this paper, we propose a personalization approach for LLMs using RAG and KGs.
Our approach focuses on using smaller models to achieve efficient results, which can be further deployed on personal devices such as smartphones. Our personalization approach targets the understanding and generation of personal results such as smart reply~\cite{kannan2016smart}. The on-device applications that store and contain user information like calendars, conversational chats, and emails can be structured into KGs, which are used for smart-response generation by the language models.
Since KGs are dynamically updated, the LLMs will have a factually correct smart response to the queries.
Multiple approaches optimize LLM model sizes and executions on edge-computing devices and smartphones.
We also make our source code\footnote{
\textbf{GitHub repository}: \url{https://github.com/asu-kim/personal-llm-kg}
}$^{,}$\footnote{
\textbf{Source code released as DOI}: \url{https://doi.org/10.5281/zenodo.14873185}
} and dataset\footnote{
\textbf{Hugging Face}: \url{https://huggingface.co/datasets/asu-kim/conversation-calendar}}$^{,}$\footnote{
\textbf{Dataset DOI}: \url{https://doi.org/10.57967/hf/4500}
} \underline{publicly available} for those who want to evaluate relevant personalization approaches.


Our proposed approach can improve the LLM inference latency while better protecting the privacy of sensitive user data locally stored on user devices, compared to the cloud-based LLMs.
We acknowledge that some on-device applications already share or sync the local private data with the application-specific cloud servers, for example, Gmail with Google, Apple Calendar with iCloud, etc.
We do not claim that our approach protects user privacy against such application-specific clouds.
Rather, we prevent the sensitive user data from being combined and sent to the cloud-based LLM provider for personalization, which can put user privacy at greater risk because the LLM provider can see all combined user data.

%
\begin{figure}
    \centering
    \includegraphics[width=0.85\columnwidth]{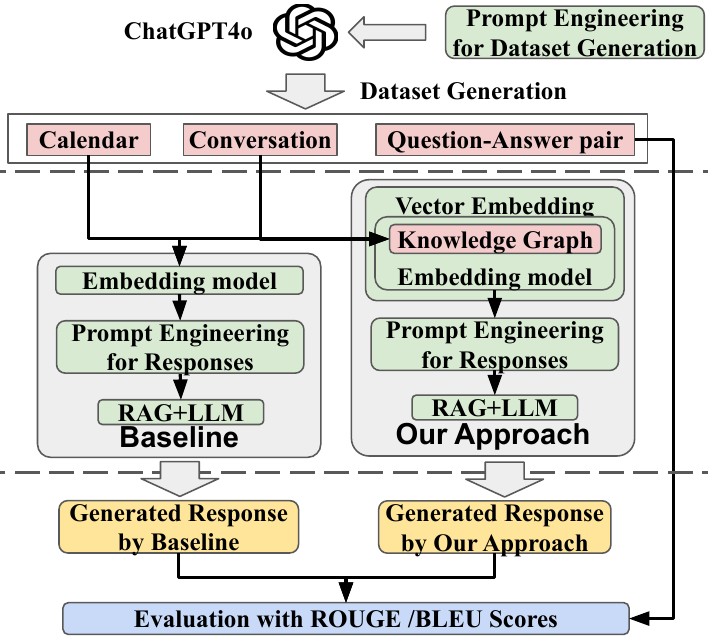}
    \caption{Overall workflow of the proposed approach}
    \label{fig:flowchart}
    \vspace{-10pt}
\end{figure}
\section{Proposed Approach}
We propose to leverage the integration of LLMs and KGs during the inference stage and generate personalized responses. Personal sensitive information like conversational data or calendar data is stored in the form of a KG. 
We consider open-source models like Llama-2-Chat (7B, 13B, and 70B)~\cite{touvron2023llama} for the evaluation in our approach, the reason being Llama 2-Chat is tailored for dialogue use cases.
Due to their reduced parameter count and streamlined architecture, these models exhibit significantly accelerated training and inference speeds.

\paragraph{\textbf{Overall Workflow}}
\figurename~\ref{fig:flowchart} shows the overall workflow of our systematic approach to be detailed in this section.
The overall workflow of this approach begins with dataset generation. Using prompt engineering, we generate unbiased data that includes calendar and conversational datasets. Additionally, we create question-and-golden-answer\footnote{We provide a detailed explanation of the golden answer in Section~\ref{sec:eval}.} pairs as part of our dataset with the help of ChatGPT. 
We use our dataset as input for the baseline model. For our proposed approach, we build a KG based on the generated dataset. Next, we utilize an embedding model to convert the input into embeddings that are essential for sequence matching during prompt-based response generation.
We then fine-tune the prompts for the retrieval pipeline and response generation. Finally, we evaluate the answers generated by both our approach and the baseline using ROUGE~\cite{lin2004rouge} and BLEU score metrics~\cite{papineni2002bleu}.

\paragraph{\textbf{Dataset Generation}}
We use the ChatGPT 4o model to generate a calendar dataset and conversational data that considers the details of information from the calendar to suit real-time conversations.
The raw data generated from the model are created in JSON and text format for readability and easy understanding. The raw data is given to the RAG-based model as a context to the baseline model.

\begin{figure}
    \centering
    \includegraphics[width=1.0\columnwidth]{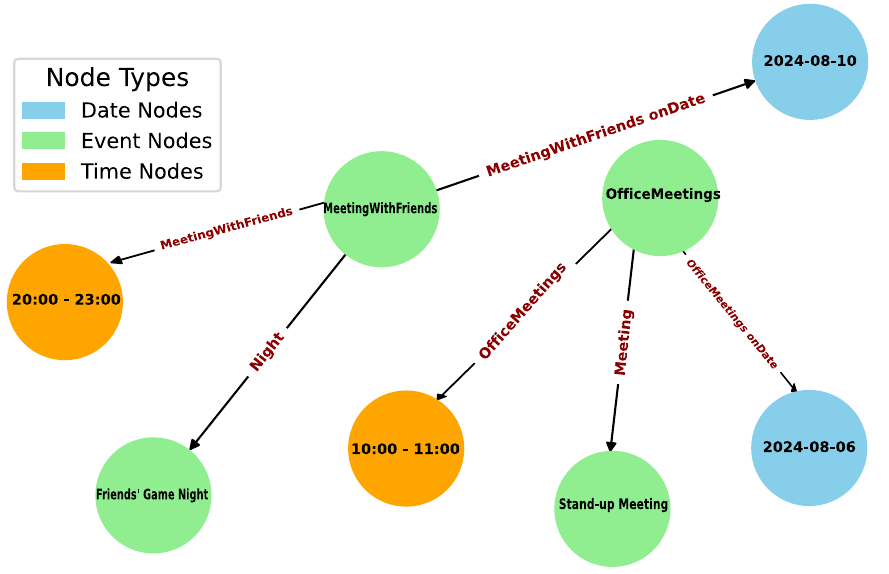}
    \caption{Knowledge graph for a sample of the calendar data from our dataset.}
    \label{fig:calendar_knowledge_graph}
    \vspace{-10pt}
\end{figure}

\paragraph{\textbf{Knowledge Graphs}}
The knowledge graph contains sources, targets, and edges represented as a set of triples. We first extract the triples from the generated calendar and conversation data, using the SpaCy library for relation extraction. This step parses the different parts of the English sentence and chooses the root verb as the edge, the subject words as the source, and the object words as the target, as seen in the \figurename~\ref{fig:calendar_knowledge_graph}. We have used the Networkx~\cite{hagberg2008exploring} Python library that is also used to create the visualization of the KG since it is used for network analysis. This dataset is given to the RAG-based model as a context to the model that forms our approach.

\paragraph{\textbf{Response Generation}}
We convert our KG and generated data (calendar and conversation) into vector embeddings using a pre-trained HuggingFace embedding model (sentence-transformers/ paraphrase-multilingual-MiniLM-L12-v2~\cite{reimers-2019-sentence-bert}). The program loads a pre-trained Llama-2-Chat models (7B\footnote{\url{https://huggingface.co/meta-llama/Llama-2-7b-chat-hf}}, 13B\footnote{\url{https://huggingface.co/meta-llama/Llama-2-13b-chat-hf}}, and 70B\footnote{\url{https://huggingface.co/meta-llama/Llama-2-70b-chat-hf}}), models from HuggingFace for text generation. We apply bitsandbytes\footnote{https://huggingface.co/bitsandbytes} quantization to optimize memory usage. The parameters like token limit, repetition penalty, and temperature are tweaked to achieve the best results. We use FAISS~\cite{douze2024faiss} for vector store, which leverages document embeddings to enable efficient similarity-based retrieval of relevant documents. 
\begin{compactItemize}
\item \textbf{Prompt engineering:}
To guide the generation of responses, we define Prompt Templates to get a concise response. To achieve the best results from each model, we experimented with different prompts by varying the types of sentences and words used. The prompt template used is:
\begin{displayquote}\texttt{Retrieve the answer from the knowledge graph <context> and generate a concise response to the <query>
}
\end{displayquote}


\item \textbf{Retrieval Pipeline:}

We use a RetrievalQA pipeline imported from LangChain\footnote{\url{https://www.langchain.com/}}, which is defined separately for the two compared approaches.
We create two separate vector spaces for the baseline and our approach.
For the baseline, we feed the generated calendar and conversation data directly to the embedding model~\cite{reimers-2019-sentence-bert}.
For our approach, we feed the KG to the embedding model.
The retriever fetches the top-k most relevant documents based on similarity to the query.
Based on the query and context obtained from the similarity search we obtain a response from the Llama-2-Chat models. 
\end{compactItemize}
Then, we compare the generated responses against the answers in the question-answer pairs to compute evaluation scores.

\section{Dataset Generation}



For evaluation, we searched for a dataset with a thorough compilation of an individual's information, including a collection of questions and answers.
To the best of our knowledge, we could not find such a dataset even after exploring open-source datasets available on platforms such as Hugging Face, Kaggle, and other online repositories.
Thus, we generate our dataset systematically and unbiased as follows.
Firstly, we use the cloud-based state-of-the-art large language model, ChatGPT-4, to generate calendar data for an individual named `Alex' in JSON format.
We utilize ChatGPT-4 for the creation of our dataset since generative models excel at producing human-like text.
We intend to match the conversational data with calendar events. This is why we use ChatGPT-4 instead of generating a calendar randomly (e.g., programmatically using Python). The events created on the calendar are more human-like and not biased.


Our dataset includes various personal events in the form of calendar data on diverse topics like family events, office meetings, and meetings with friends.
Secondly, our dataset is also combined with conversations between Alex and his friends, family, AI assistants, acquaintances, etc., about entertainment, games, Business startup plans, customer care support, health and wellness, and group trip planning.
Each conversation in our dataset is a single session, with the length being 10 to 20 messages. 
The prompt used to generate calendar data is:

\begin{displayquote}\texttt{Generate a calendar dataset for a person named Alex having events such as family events, office meetings, holidays, and meetings with friends for August, September, October, November, and December.
}
\end{displayquote}
\noindent
To create the question and golden answer pair for the evaluation of our method, we prompt the ChatGPT 4o as follows:

\begin{displayquote}\texttt{Generate a set of <number> questions with exact answers based on the calendar data.
}
\end{displayquote}
The prompt used to generate conversation data is: 
\begin{displayquote}\texttt{Generate a conversation with 20 messages like Text messages, WhatsApp, or messenger messages for <topic>.
Main person: Alex 
Friends : <names>
Use the calendar data to see Alex's schedule.
}
\end{displayquote}
To create the question and golden answer pair for the evaluation of our method, we prompt the ChatGPT 4o as:

\begin{displayquote}\texttt{Generate a set of <number> questions with exact answers based on the conversation.
}
\end{displayquote}

Listing~\ref{lst:calendar_data} shows an example of calendar items from our dataset.
Our dataset is publicly available at Hugging Face.\footnote{\url{https://huggingface.co/datasets/asu-kim/conversation-calendar}}

\begin{lstlisting}[basicstyle=\ttfamily\footnotesize,
caption=Example calendar data in JSON format.,
breaklines=true,
breakatwhitespace=true,
label=lst:calendar_data]
"AlexCalendar2024": {
  "January": [
    { "event": "Catch-up with Friends", "date": "2024-01-10", "time": "16:00 - 17:30" },
    { "event": "Team Meeting",         "date": "2024-01-15", "time": "09:00 - 10:00" },
    { "event": "Family Dinner",        "date": "2024-01-20", "time": "18:00 - 20:00" }
  ]}
\end{lstlisting}

\section{Evaluation}
\label{sec:eval}

\begin{figure*}
\centering
    \begin{subfigure}{0.32\linewidth}
        \centering
        \includegraphics[width=1.0\columnwidth]{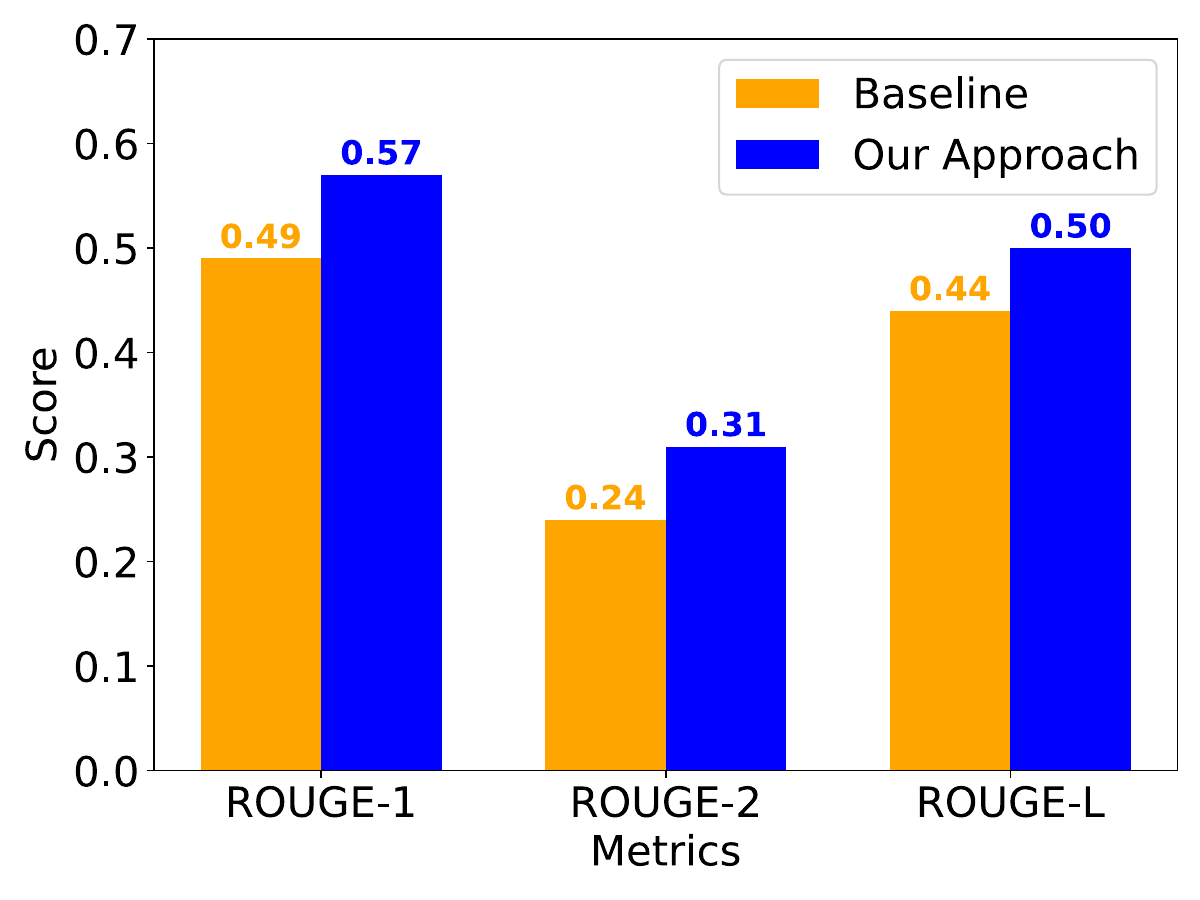}
        \caption{7B ROUGE Scores}
        \label{fig:Average_Rouge_Score_7B}
    \end{subfigure}
    \hfill
    \begin{subfigure}{0.32\linewidth}
        \centering
        \includegraphics[width=1.0\columnwidth]{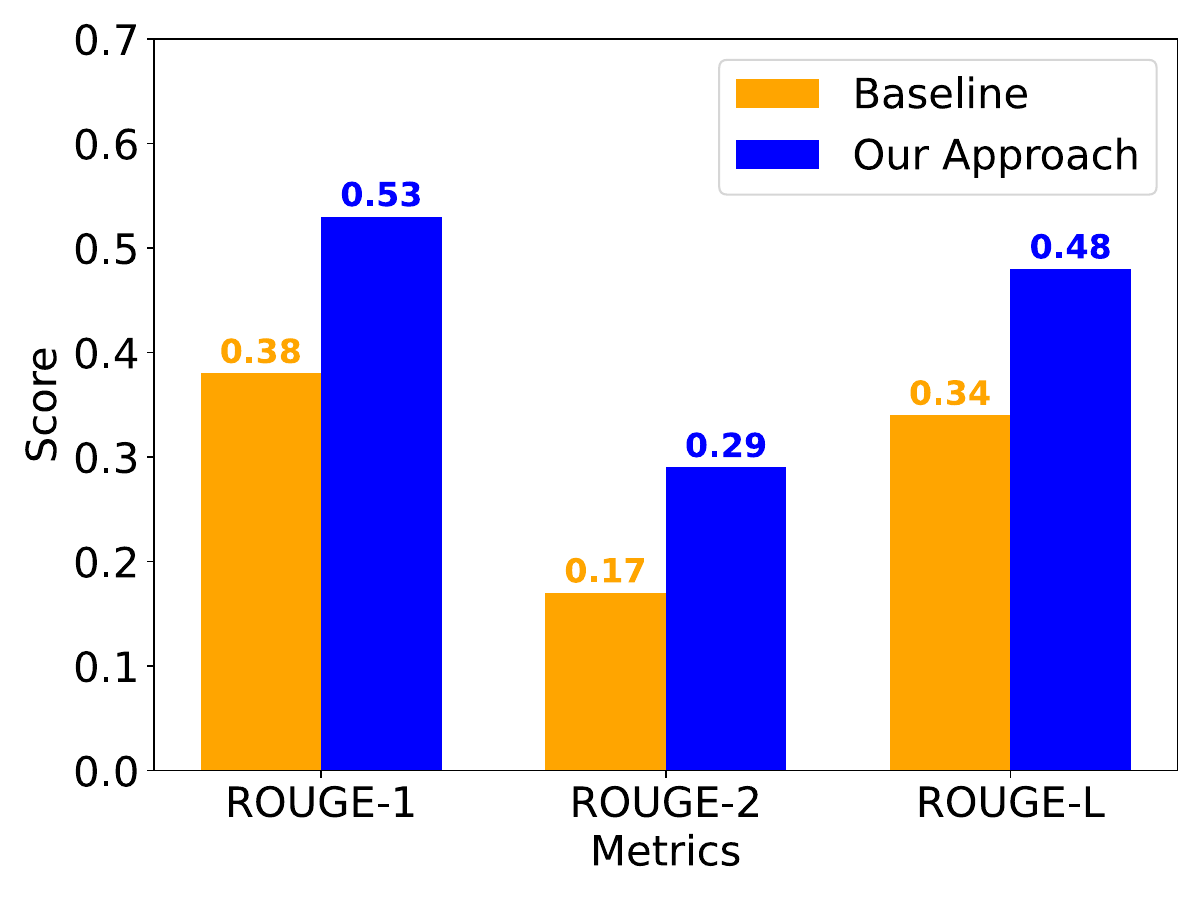}
        \caption{13B ROUGE Scores}
        \label{fig:Average_Rouge_Score_13B}
    \end{subfigure}
    \hfill
    \begin{subfigure}{0.32\linewidth}
        \centering
        \includegraphics[width=1.0\columnwidth]{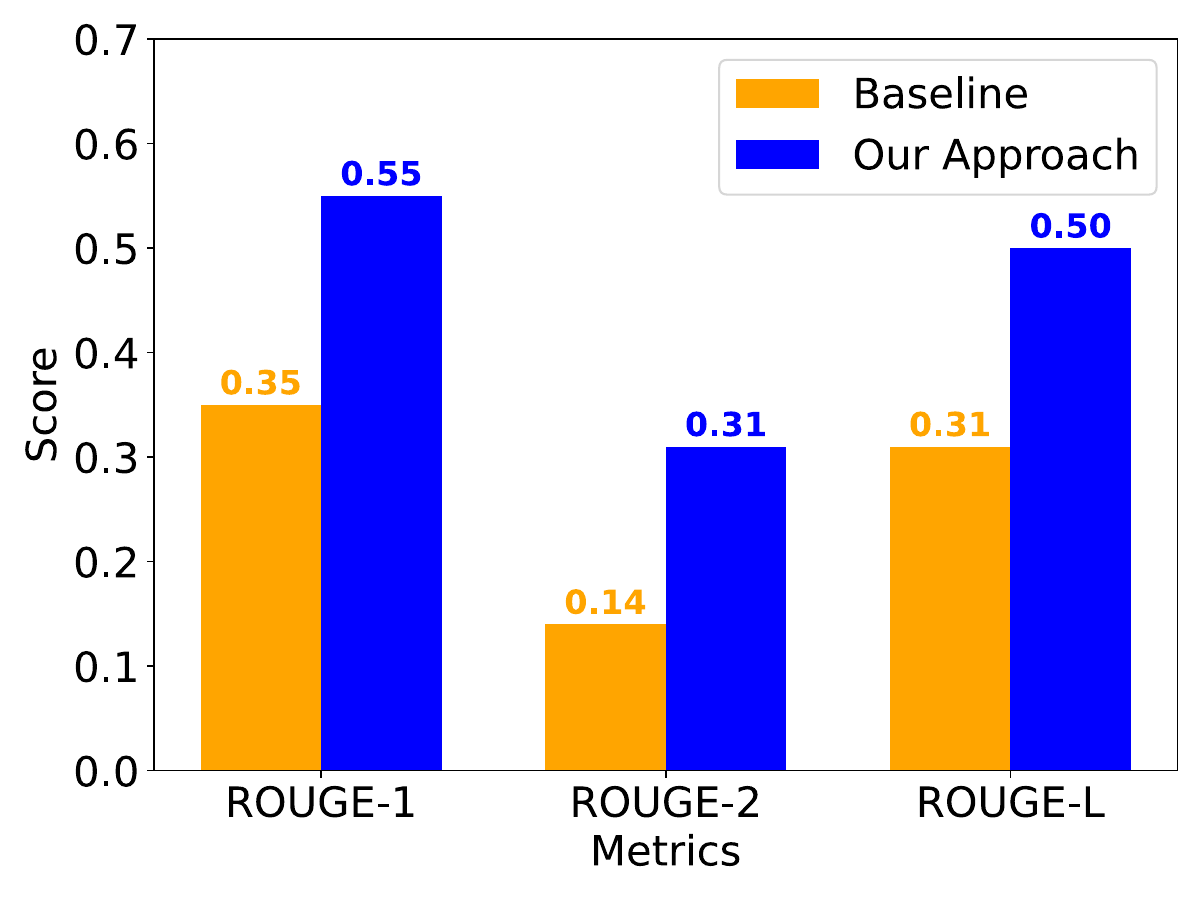}
        \caption{70B ROUGE Scores}
        \label{fig:Average_Rouge_Score_70B}
    \end{subfigure}
    \caption{Experimental results comparing three types of ROUGE metric scores (ROUGE-1, ROUGE-2, and ROUGE-L) of three Llama-2-Chat models (7B, 13B, and 70B) between the baseline and our approach.}
    \label{fig:horizontal_results}
\end{figure*}
\begin{figure}
    \centering
    \includegraphics[width=0.7\columnwidth]{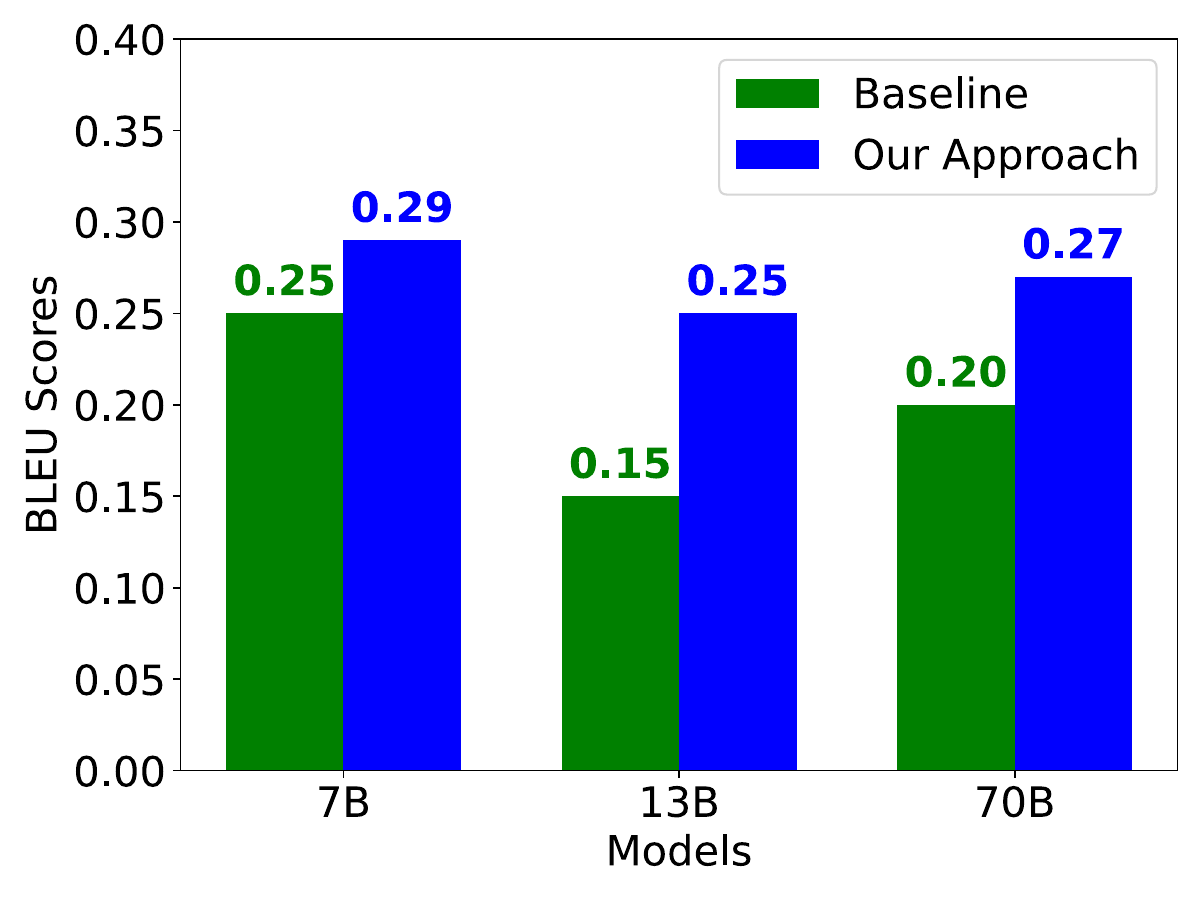}
    \caption{Experimental results comparing BLEU scores of Llama-2-Chat models (7B, 13B, and 70B) between the baseline and our approach.}
    \label{fig:BLEU_score}
\end{figure}
Our evaluation focuses on how well the model can use the information from the knowledge graph (KG) and generate responses. 
The evaluation metrics include ROUGE (precision, recall, and F1-score), BLEU score, and execution time, ensuring a comprehensive comparison between the baseline and the proposed approach. 
The evaluation also compares the execution time for each query. In the upcoming sections, we present our experimental setup, dataset, baseline, and experimental results. 


\paragraph{\textbf{Baseline}}
The traditional RAG~\cite{10.5555/3495724.3496517} achieved state-of-the-art performance in various information retrieval and question-answering tasks~\cite{fan2024survey}~\cite{10.1145/3626772.3661370}. 
We evaluate our model by choosing the RAG architecture as a baseline model because it fetches the relevant documents and generates accurate responses.
The RAG implementation is imported from LangChain, and it is integrated with our dataset.

\paragraph{\textbf{Experimental Setup}}
We conduct experiments on a workstation with an Intel i9 processor and an NVIDIA RTX A6000 GPU.
For the embedding model, we utilize the sentence transformer paraphrase-multilingual-MiniLM-L12-v2.
The pre-trained model deployed for inference is available in the Hugging Face repository.
GPU performance and memory utilization are monitored using NVIDIA-SMI, with the environment configured to use CUDA Version 12.4.

\paragraph{\textbf{Evaluation Criteria}}
We use the \emph{golden answer} as the ground truth for evaluation, which is generated during the dataset creation stage without any human bias. The responses from our approach and the baseline are evaluated against golden answers. The ROUGE score is used as an evaluation metric for unigram, bigram, and longest sequence matching. The BLEU score measures the n-gram similarity between the golden answer and the generated response.


\begin{table}[]
\begin{tabular}{|l|}
\hline
\textbf{Question 8 - Query:} \\ 
What is the event on August 19th, 2024? \\ \hline
\textbf{Golden Answer:} \\ 
The event on 2024-08-19 is “Raksha Bandhan,” observed all day. \\ \hline
\textbf{Generated Response by Baseline:} \\ 
\begin{tabular}[c]{@{}l@{}}Please do not provide any unnecessary information.\\    Additional Notes:I want to retrieve only the date and time of\\ the event on August 19th, 2024.\\       Please help me with this.\\     Thank you!\end{tabular} \\ \hline
\textbf{Generated Response by Our Approach:} \\ 
\begin{tabular}[c]{@{}l@{}}Based on the given text, the event on August 19th, 2024 is\\ Raksha Bandhan.\end{tabular} \\ \hline
\end{tabular}
\caption{Sample responses by the baseline and our approach.}
\label{tab:sample_resp}
\vspace{-15pt}
\end{table}

\paragraph{\textbf{Experimental Results}}
We measure three types of ROUGE scores and the BLEU score, which map the n-gram overlap of words.
The generation of the correct knowledge graph plays an important role in the generation of our results. 
During the query process, the model can easily find similar words without losing its way, and hence, the KGs serve as a roadmap to the right solution, thereby reducing hallucinations by the LLM. Since our approach can direct the LLM to the right solution, the response times are better. 
Here are graphs showing experimental results using ROUGE scores and BLEU scores on the baseline vs. our approach. Our approach outperforms the baseline across all three Llama-2-Chat models (7B, 13B, and 70B), as shown for ROUGE in \figurename~\ref{fig:Average_Rouge_Score_7B}, \figurename~\ref{fig:Average_Rouge_Score_13B}, and \figurename~\ref{fig:Average_Rouge_Score_70B} and for BLEU-1 in \figurename~\ref{fig:BLEU_score}. The response shows that the model is being directed to the right data and is generating the required response, as seen in the sample results shown in \tablename~\ref{tab:sample_resp}.


    
    
    

Our results in \figurename\text{s}~\ref{fig:horizontal_results} and~\ref{fig:BLEU_score} and \tablename~\ref{tab:sample_resp} demonstrate that the generated response for the same model with the knowledge graph (our approach) leads to a better response closer to the golden answer as it is directed toward the right context.
Additionally, the average execution time for 20 question-response pairs shows that our approach is faster, as seen in \tablename~\ref{tab:execution-times}.



\begin{table}[]
\begin{tabular}{|c|r|rr|}
\hline
\multicolumn{1}{|l|}{\multirow{2}{*}{}} & \multicolumn{1}{l|}{}                    & \multicolumn{2}{c|}{\textbf{Execution time (seconds)}}                              \\ \cline{2-4} 
\multicolumn{1}{|l|}{\textbf{LLM Model}}                  & \multicolumn{1}{c|}{\textbf{Parameters}} & \multicolumn{1}{l|}{\textbf{Baseline}} & \multicolumn{1}{c|}{\textbf{Our Approach}} \\ \hline
\textbf{Llama-2-Chat}                   & 7B                                       & \multicolumn{1}{r|}{0.81}              & 0.70                                       \\ \hline
\textbf{Llama-2-Chat}                   & 13B                                      & \multicolumn{1}{r|}{1.20}              & 1.00                                       \\ \hline
\textbf{Llama-2-Chat}                   & 70B                                      & \multicolumn{1}{r|}{3.70}              & 3.50                                       \\ \hline
\end{tabular}
\caption{Execution times of different Llama-2-Chat models under the baseline and our approach.}
\label{tab:execution-times}
\vspace{-15pt}
\end{table}

\begin{table*}[]
\begin{tabular}{|
>{\columncolor[HTML]{FFFFFF}}r 
>{\columncolor[HTML]{FFFFFF}}r |
>{\columncolor[HTML]{FFFFFF}}r 
>{\columncolor[HTML]{FFFFFF}}r 
>{\columncolor[HTML]{FFFFFF}}r |
>{\columncolor[HTML]{FFFFFF}}r 
>{\columncolor[HTML]{FFFFFF}}r 
>{\columncolor[HTML]{FFFFFF}}r |
>{\columncolor[HTML]{FFFFFF}}r 
>{\columncolor[HTML]{FFFFFF}}r
>{\columncolor[HTML]{FFFFFF}}r|}\hline
\multicolumn{2}{|r|}{\cellcolor[HTML]{FFFFFF}\textbf{Metrics}}                                       & \multicolumn{3}{c|}{\cellcolor[HTML]{FFFFFF}\textbf{ROUGE-1}}                                                                                                                             & \multicolumn{3}{c|}{\cellcolor[HTML]{FFFFFF}\textbf{ROUGE-2}}                                                                                                                             & \multicolumn{3}{c|}{\cellcolor[HTML]{FFFFFF}\textbf{ROUGE-L}}                                                                                                                             \\ \hline
\multicolumn{2}{|l|}{\cellcolor[HTML]{FFFFFF}\textbf{Llama Models}}                                  & \multicolumn{1}{c|}{\cellcolor[HTML]{FFFFFF}\textbf{Precision}} & \multicolumn{1}{c|}{\cellcolor[HTML]{FFFFFF}\textbf{Recall}} & \multicolumn{1}{c|}{\cellcolor[HTML]{FFFFFF}\textbf{F1}} & \multicolumn{1}{c|}{\cellcolor[HTML]{FFFFFF}\textbf{Precision}} & \multicolumn{1}{c|}{\cellcolor[HTML]{FFFFFF}\textbf{Recall}} & \multicolumn{1}{c|}{\cellcolor[HTML]{FFFFFF}\textbf{F1}} & \multicolumn{1}{c|}{\cellcolor[HTML]{FFFFFF}\textbf{Precision}} & \multicolumn{1}{c|}{\cellcolor[HTML]{FFFFFF}\textbf{Recall}} & \multicolumn{1}{c|}{\cellcolor[HTML]{FFFFFF}\textbf{F1}} \\ \hline
\multicolumn{1}{|r|}{\cellcolor[HTML]{FFFFFF}}                               & \textbf{Baseline}     & \multicolumn{1}{r|}{\cellcolor[HTML]{FFFFFF}0.423}              & \multicolumn{1}{r|}{\cellcolor[HTML]{FFFFFF}0.599}           & 0.490                                                    & \multicolumn{1}{r|}{\cellcolor[HTML]{FFFFFF}0.206}              & \multicolumn{1}{r|}{\cellcolor[HTML]{FFFFFF}0.297}           & 0.241                                                    & \multicolumn{1}{r|}{\cellcolor[HTML]{FFFFFF}0.385}              & \multicolumn{1}{r|}{\cellcolor[HTML]{FFFFFF}0.543}           & 0.445                                                    \\ \cline{2-11} 
\multicolumn{1}{|r|}{\multirow{-2}{*}{\cellcolor[HTML]{FFFFFF}\textbf{7B}}}  & \textbf{Our Approach} & \multicolumn{1}{r|}{\cellcolor[HTML]{FFFFFF}0.549}              & \multicolumn{1}{r|}{\cellcolor[HTML]{FFFFFF}0.619}           & 0.567                                                    & \multicolumn{1}{r|}{\cellcolor[HTML]{FFFFFF}0.296}              & \multicolumn{1}{r|}{\cellcolor[HTML]{FFFFFF}0.343}           & 0.311                                                    & \multicolumn{1}{r|}{\cellcolor[HTML]{FFFFFF}0.485}              & \multicolumn{1}{r|}{\cellcolor[HTML]{FFFFFF}0.548}           & 0.502                                                    \\ \hline
\multicolumn{1}{|r|}{\cellcolor[HTML]{FFFFFF}}                               & \textbf{Baseline}     & \multicolumn{1}{r|}{\cellcolor[HTML]{FFFFFF}0.325}              & \multicolumn{1}{r|}{\cellcolor[HTML]{FFFFFF}0.464}           & 0.375                                                    & \multicolumn{1}{r|}{\cellcolor[HTML]{FFFFFF}0.145}              & \multicolumn{1}{r|}{\cellcolor[HTML]{FFFFFF}0.204}           & 0.166                                                    & \multicolumn{1}{r|}{\cellcolor[HTML]{FFFFFF}0.297}              & \multicolumn{1}{r|}{\cellcolor[HTML]{FFFFFF}0.416}           & 0.340                                                    \\ \cline{2-11} 
\multicolumn{1}{|r|}{\multirow{-2}{*}{\cellcolor[HTML]{FFFFFF}\textbf{13B}}} & \textbf{Our Approach} & \multicolumn{1}{r|}{\cellcolor[HTML]{FFFFFF}0.493}              & \multicolumn{1}{r|}{\cellcolor[HTML]{FFFFFF}0.597}           & 0.530                                                    & \multicolumn{1}{r|}{\cellcolor[HTML]{FFFFFF}0.272}              & \multicolumn{1}{r|}{\cellcolor[HTML]{FFFFFF}0.315}           & 0.286                                                    & \multicolumn{1}{r|}{\cellcolor[HTML]{FFFFFF}0.449}              & \multicolumn{1}{r|}{\cellcolor[HTML]{FFFFFF}0.537}           & 0.480                                                    \\ \hline
\multicolumn{1}{|r|}{\cellcolor[HTML]{FFFFFF}}                               & \textbf{Baseline}     & \multicolumn{1}{r|}{\cellcolor[HTML]{FFFFFF}0.333}              & \multicolumn{1}{r|}{\cellcolor[HTML]{FFFFFF}0.410}           & 0.354                                                    & \multicolumn{1}{r|}{\cellcolor[HTML]{FFFFFF}0.129}              & \multicolumn{1}{r|}{\cellcolor[HTML]{FFFFFF}0.156}           & 0.137                                                    & \multicolumn{1}{r|}{\cellcolor[HTML]{FFFFFF}0.294}              & \multicolumn{1}{r|}{\cellcolor[HTML]{FFFFFF}0.360}           & 0.312                                                    \\ \cline{2-11} 
\multicolumn{1}{|r|}{\multirow{-2}{*}{\cellcolor[HTML]{FFFFFF}\textbf{70B}}} & \textbf{Our Approach} & \multicolumn{1}{r|}{\cellcolor[HTML]{FFFFFF}0.573}              & \multicolumn{1}{r|}{\cellcolor[HTML]{FFFFFF}0.566}           & 0.555                                                    & \multicolumn{1}{r|}{\cellcolor[HTML]{FFFFFF}0.328}              & \multicolumn{1}{r|}{\cellcolor[HTML]{FFFFFF}0.307}           & 0.309                                                    & \multicolumn{1}{r|}{\cellcolor[HTML]{FFFFFF}0.517}              & \multicolumn{1}{r|}{\cellcolor[HTML]{FFFFFF}0.507}           & 0.498                                                    \\ \hline
\end{tabular}
\caption{Precision, Recall, and F1 scores for ROUGE-1, ROUGE-2, and ROUGE-L across 7B, 13B, and 70B models}

\label{tab:Precision, Recall and F1}
\end{table*}
Our approach achieves significant improvements across all evaluation metrics. We observe from \tablename~\ref{tab:Precision, Recall and F1} that our approach outperforms the baseline in precision, recall, and F1 scores of ROUGE-1 (unigram), ROUGE-2 (bigram), ROUGE-L (longest sequence overlap). We observed an average increase of 35.15\% in ROUGE-1, 65.57\% in ROUGE-2, and 35.82\% in ROUGE-L scores. For BLEU-1, the average increase was 61.11\%. Additionally, the execution time showed an overall reduction of 8.931\%. These improvements indicate that our approach enhances text generation quality and increases lexical overlap with the golden answer. 

\section{Conclusion}
In summary, utilizing KGs enables enhanced adaptation for specific domains, guaranteeing the model produces responses customized to the particular context (e.g., calendar dataset). Our approach can improve the execution time, enabling high-quality smart responses using smaller LLMs that can run on the local device. This approach forms the basis for on-device LLM, where we have continuously changing data. The knowledge database can be utilized to store and update the data on time. This reduces the risk of sending personal information to the cloud. We plan on increasing the ability of smaller models to perform well on unseen data. 
Building a knowledge graph eliminates the need to send sensitive data to the LLM provider.
\begin{acks}
This work was supported in part by the NSF I/UCRC for Intelligent, Distributed, Embedded
Applications and Systems (IDEAS) and from NSF grant \#2231620.
This work was supported in part by ATTO Research Co., Ltd.
\end{acks}

\bibliographystyle{ACM-Reference-Format}
\bibliography{sample-base}

\end{document}